# A Statistical Approach to Adult Census Income Level Prediction

Navoneel Chakrabarty
Computer Science and Engineering ,
Jalpaiguri Government Engineering College ,
Jalpaiguri, West Bengal, India
Email: nc2012@cse.jgec.ac.in

Sanket Biswas
Computer Science and Engineering ,
Jalpaiguri Government Engineering College,
Jalpaiguri, West Bengal, India
Email: sanket1824@cse.jgec.ac.in

*Abstract*—The prominent inequality of wealth and income is a huge concern especially in the United States. The likelihood of diminishing poverty is one valid reason to reduce the world's surging level of economic inequality. The principle of universal moral equality ensures sustainable development and improve the economic stability of a nation. Governments in different countries have been trying their best to address this problem and provide an optimal solution. This study aims to show the usage of machine learning and data mining techniques in providing a solution to the income equality problem. The UCI Adult Dataset has been used for the purpose. Classification has been done to predict whether a person's yearly income in US falls in the income category of either greater than 50K Dollars or less equal to 50K Dollars category based on a certain set of attributes. The Gradient Boosting Classifier Model was deployed which clocked the highest accuracy of 88.16%, eventually breaking the benchmark accuracy of existing works.

## I. INTRODUCTION

Over the last two decades, humans have grown a lot of dependence on data and information in society and with this advent growth, technologies have evolved for their storage, analysis and processing on a huge scale. The fields of Data Mining and Machine Learning have not only exploited them for knowledge and discovery but also to explore certain hidden patterns and concepts which led to the prediction of future events, not easy to obtain. The problem of income inequality has been of great concern in the recent years. Making the poor better off does not seem to be the sole criteria to be in quest for eradicating this issue. People of the United States believe that the advent of economic inequality is unacceptable and demands a fair share of wealth in the society. This model actually aims to conduct a comprehensive analysis to highlight the key factors that are necessary in improving an individual's income. Such an analysis helps to set focus on the important areas which can significantly improve the income levels of individuals.

This paper has been structured as an introduction, literature review, proposed methodology, training the model, implementation details, results and conclusion

## II. LITERATURE REVIEW

Certain efforts using machine learning models have been made in the past by researchers for predicting income levels.

- Chockalingam et. al. [1] explored and analysed the Adult Dataset and used several Machine Learning Models like Logistic Regression, Stepwise Logistic Regression, Naive Bayes, Decision Trees, Extra Trees, k-Nearest Neighbor, SVM, Gradient Boosting and 6 configurations of Activated Neural Network. They also drew a comparative analysis of their predictive performances.
- Bekena [2] implemented the Random Forest Classifier algorithm to predict income levels of individuals.
- Topiwalla [3] made the usage of complex algorithms like XGBOOST, Random Forest and stacking of models for prediction tasks including Logistic Stack on XGBOOST and SVM Stack on Logistic for scaling up the accuracy.
- Lazar [4] implemented Principal Component Analysis (PCA) and Support Vector Machine methods to generate and evaluate income prediction data based on the Current Population Survey provided by the U.S. Census Bureau.
- Deepajothi et. al. [5] tried to replicate Bayesian Networks, Decision Tree Induction, Lazy Classifier and Rule Based Learning Techniques for the Adult Dataset and presented a comparative analysis of the predictive performances.
- Lemon et. al. [6] attempted to identify the important features in the data that could help to optimize the complexity of different machine learning models used in classification tasks.
- Haojun Zhu [7] attempted Logistic Regression as the Statistical Modelling Tool and 4 different Machine Learning Techniques, Neural Network, Classification and Regression Tree, Random Forest, and Support Vector Machine for predicting Income Levels.

## III. PROPOSED METHODOLOGY

### A. The Dataset

The data for our study was accessed from the University of California Irvine (UCI) Machine Learning Repository [8]. It was actually extracted by Barry Becker using the 1994 census database. The data set includes figures on 48,842 different records and 14 attributes for 42 nations. The 14 attributes consist of 8 categorical and 6 continuous attributes containing information on age, education, nationality, marital status, relationship status, occupation, work classification, gender, race, working hours per week, capital loss and capital gain as shown in Table 1. The binomial label in the data set is the income level which predicts whether a person earns more than 50 Thousand Dollars per year or not based on the given set of attributes.

TABLE I
FEATURES AND EXTRA TREES SCORE

| ID | Attribute Name | Extra Trees Classifier Score |
|---|---|---|
| F1 | age (continuous) | 0.1659 |
| F2 | workclass (categorical) | 0.0484 |
| F3 | fnlwgt (continuous) | 0.1636 |
| F4 | education (categorical) | 0.0329 |
| F5 | education-num (continuous) | 0.0901 |
| F6 | marital-status (categorical) | 0.0604 |
| F7 | occupation (categorical) | 0.0773 |
| F8 | relationship (categorical) | 0.0966 |
| F9 | race (categorical) | 0.0144 |
| F10 | sex (categiorical) | 0.0243 |
| F11 | capital-gain (continuous) | 0.0847 |
| F12 | capital-loss (continuous) | 0.0257 |
| F13 | hours-per-week (continuous) | 0.0978 |
| F14 | native-country (categorical) | 0.0177 |

### B. Feature Study and Selection

Based on the scores of the Extra Tree Classifier for different attributes (as shown from Table 1) the most relevant features have been selected, that are going to be implemented in our model. A visual explanation of the Extra Trees Classifier or Extremely Randomized Trees is shown in Fig 1. As a result Features F9 (race) and F14 (native-country) have been eliminated as they have the least Extra Trees Classifier Scores.

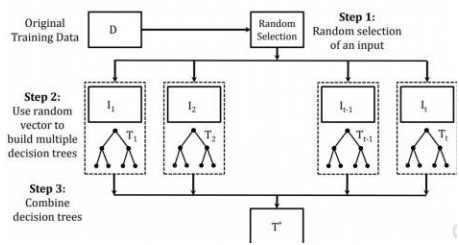

**Fig 1. Visual Explanation of Extra Trees Classifier**

A Correlation Matrix is shown in Fig 2, in the form of a Heat-Map showing Feature-to-Feature and Feature-to-Label Pearson Correlations where all the features are Continuous Variables.

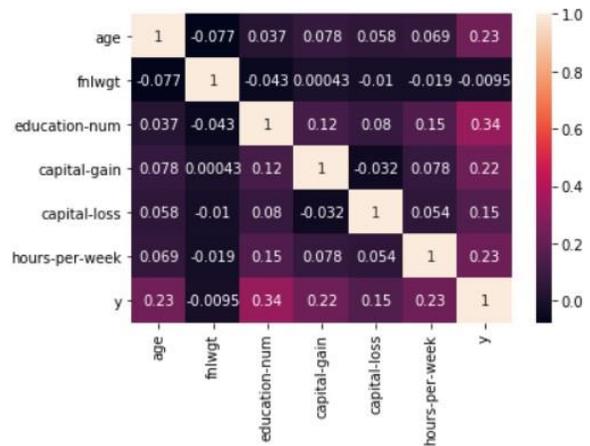

**Fig 2. Heat-Map showing Feature-to-Feature and Feature-to-Label's Pearson Correlation Coefficients**

### C. Data Visualization

Data Visualization has been done using Box and Whisker Plots of all continuous features to clearly understand the measures of their central tendencies shown in Fig 3, Fig 4, Fig 5, Fig 6, Fig 7 and Fig 8.

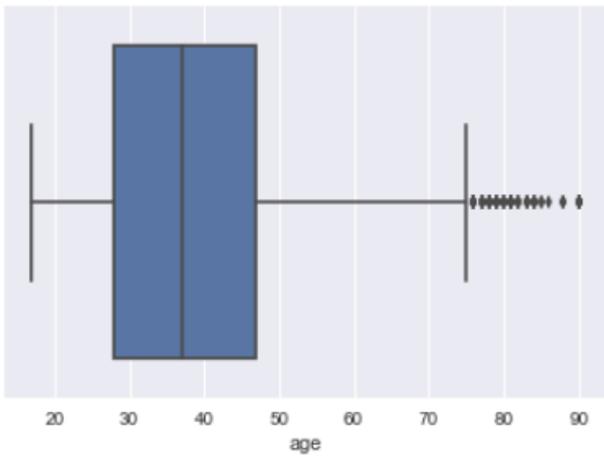

**Fig 3. Box and Whisker for 'Age' attribute**

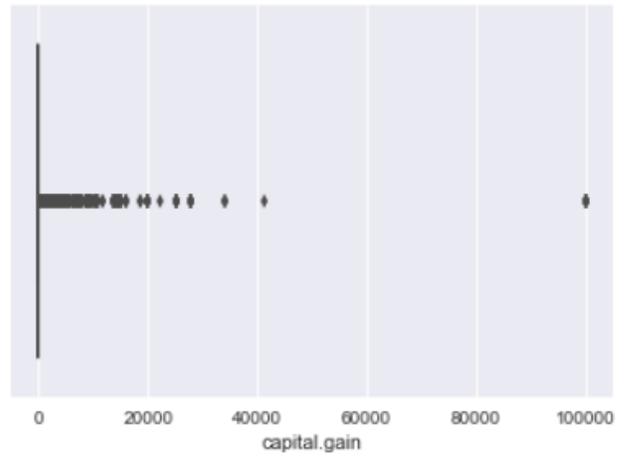

**Fig 6. Box and Whisker for 'capital.gain' attribute**

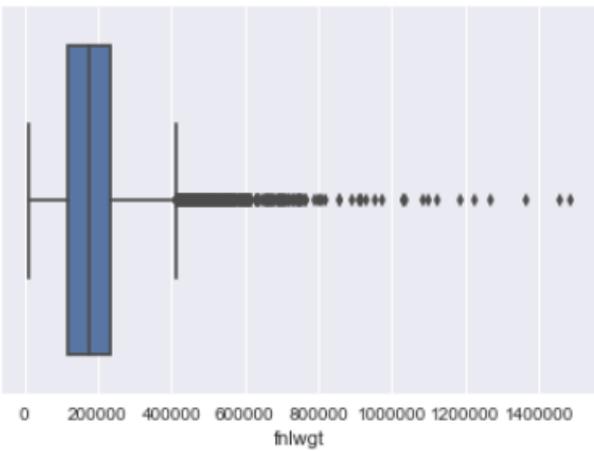

**Fig 4. Box and Whisker for 'fnlwgt' attribute**

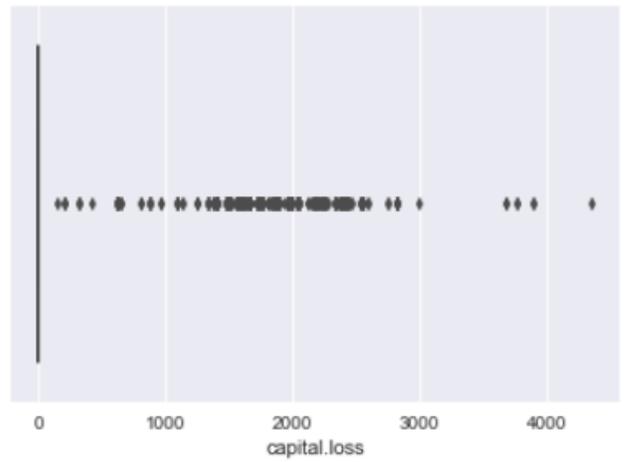

**Fig 7. Box and Whisker for 'capital.loss' attribute**

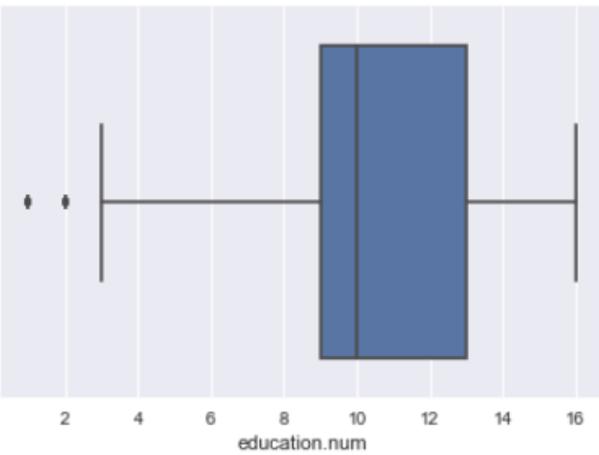

**Fig 5. Box and Whisker for 'education.num' attribute**

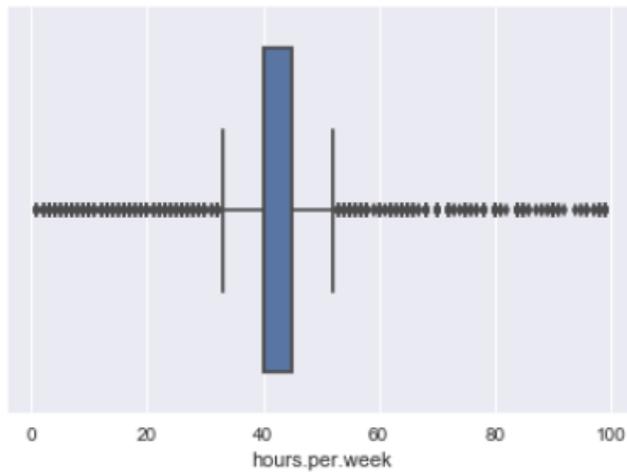

**Fig 8. Box and Whisker for 'hours.per.week' attribute**

## D. Data Preprocessing

Before processing the Adult Dataset, cleaning the data with certain preprocessing techniques becomes a necessity. This includes:

*1) Handling Missing Values:* The dataset contains certain set of missing values for categorical features, workclass, occupation, native-country which has been dealt with some algorithmic transformations applied to the data. The missing values are flexibly handled for every attribute by setting a default marker called '?' and assigning a unique category for negating information loss.

*2) Encoding of Categorical or Non-Numeric features:* As all Categorical Features are non-numeric, encoding has been done in 2 stages:

- Label Encoding: All categorical features are label encoded, where alphabetically each category is assigned numbers starting from 0. This is also done before running the Extra Trees Classifier Algorithm for efficient feature selection.
- One-Hot Encoding: This involves splitting of different categorical features into its own categories where each and every category assumes a binary value i.e., 0 if it does not belong to that category and 1 if it belongs to that category. This is important for those categorical features where there exists no ordinal relationship in between them. One-Hot Encoding has been done for categorical features having more than 2 categories. Here, for all categorical features except sex attribute, all label encoded forms are transformed into One-Hot Encoded Forms. This is because sex attribute has only 2 categories i.e., male and female, which have been already represented in binary form in a single attribute and hence to avoid the curse of dimensionality, no One-Hot Encoding is done for sex attribute.

*3) Shuffling:* The whole dataset has been shuffled in a consistent way such that all the categories of different attributes remain included in Training Set and Validation Set.

*4) Splitting:* Now, the dataset is split into training and testing sets. With 80% of the data made available for training purposes and the rest 20% is used for testing.

## E. Learning Algorithm

The learning algorithm, used to build the predictive model is an Ensemble Learning and Boosting Algorithm known as Gradient Boosting Classifier.

*1) Boosting:* It is an ensembling technique, in which predictors (which are decision trees) are being constructed sequentially rather than independently [9].

*2) Implementation:* In GBC, in the sequence of predictors being constructed, at each and every sequence the error in the previous decision tree is corrected by the decision tree following it. So, at each step, the GB Classifier, tends to fit the Training Data more and more. The Pseudo Code for Gradient Boosting Classifier is given below.

1. A learning rate alpha is assumed, say 0.1 as alpha(t)
2. A weak classifier is selected as h(t).
3. The Population Distribution is updated in the next step.

$$D_{t+1}(i) = \frac{D_t(i) \exp(-\alpha_t y_i h_t(x_i))}{Z_t}$$

$$Z_t = \sum_{i=1}^{m} D_t(i) \exp(-\alpha_t y_i h_t(x_i))$$

4. The new Population Distribution is used to construct the next learner or decision tree.
5. Steps 1-4 are iterated, until no hypothesis is found which can result in further improvement.
6. A weighted average of the frontier is taken using all the learners used till now where the weights are simply the alpha values.

$$\alpha_t = \tfrac{1}{2} \ln\left(\frac{1 - \epsilon_t}{\epsilon_t}\right)$$

**Algorithm 1:** Gradient Boosting Classifier

### IV. TRAINING THE MODEL

The Gradient Boosting Classifier Model is tuned using Grid Search Algorithm for getting the best set of hyper-parameters. After training the model with Grid-Search applied on GBC, 250 estimators and maximum depth of 4 are obtained. The summary of Grid-Search Tuning of GBC model on the basis of the Mean Score is shown in Fig 9.

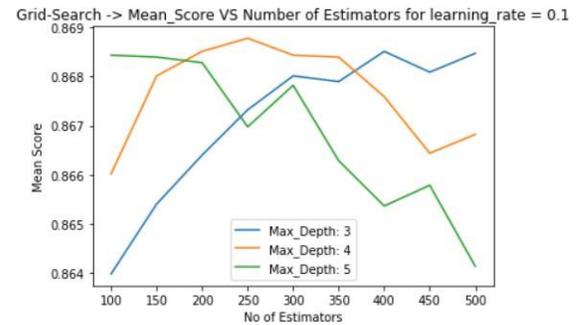

**Fig 9. Grid-Search Summary on Mean Score**

### V. IMPLEMENTATION DETAILS

The Data Preprocessing and Model Development are done using Python's Scikit-Learn Machine Learning Toolbox on a machine with Intel(R) Core(TM) i5-8250U processor, CPU @ 1.60 GHz 1.80 GHz and 8 GB RAM. The Visualizations are made using Python's Plotting Libraries, Matplotlib and Seaborn.

VI. RESULTS

Out of a total of 48,842 instances present in the dataset, 39,074 instances have been used for training while the rest 9,768 instances have been reserved for testing. After complete evaluation, the model performance are evaluated on the following metrics:

- The Training Accuracy describes the accuracy achieved on the Training Set.
  From the model, a Training Accuracy of 88.73% is achieved.
- The Validation Accuracy describes the accuracy achieved on the Validation Set.
  From the model, a Validation Accuracy of 88.16% is obtained.
- The Sensitivity or Recall is defined as the fraction of correctly identified positives.

$$Recall = TP/TP + FN$$

As a result, a Recall of 0.88 has been achieved from our model.
- Precision is defined as the proportion of correctly predicted positive observations of the total predicted positive observations.

$$Precision = TP/TP + FP$$

As a result, a Precision of 0.88 has been achieved from our model.
- F1-Score is the Harmonic Mean of Recall and Precision.

$$F_1 = \frac{2}{\frac{1}{recall} + \frac{1}{precision}} = 2 \cdot \frac{\text{precision} \cdot \text{recall}}{\text{precision} + \text{recall}}.$$

From the model, a F1-Score of 0.88 has been achieved.
- Area Under Receiver Operator Characteristic Curve (AUROC): ROC Curve is the plot of True Positive Rate vs False Positive Rate. An Area under ROC Curve of greater than 0.5, is acceptable. The respective ROC Curve is shown in Fig 12.
- Confusion Matrix has the structure shown in Fig 10:

**Fig 10. Structure of Confusion Matrix**

The Confusion Matrix for our model is shown in Fig 11.

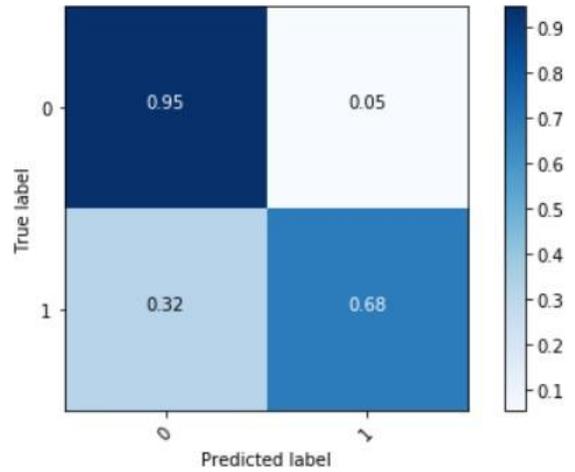

**Fig 11. Normalized Confusion Matrix for Model Performance Evaluation**

All the results are tabulated in Table 2.

| Training Accuracy | 88.73% |
|---|---|
| Validation Accuracy | 88.16% |
| Recall | 0.88 |
| Precision | 0.88 |
| F1-Score | 0.88 |
| AUROC | 0.93 |

**Table 2. Results**

The Receiver Operator Characteristic Curve (ROC Curve) is implemented using the decision function attribute of Gradient Boosting Classifier which returns values that give a measure of how far a data-point is away from the Decision Boundary from either side (negative value for opposite side). It is shown in Fig 12.

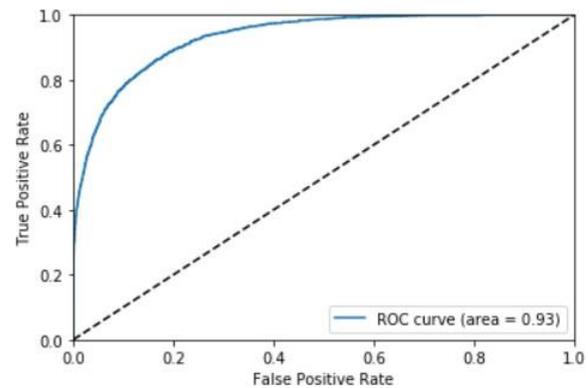

**Fig 12. ROC Curve showing the Area Under the Curve**

- According to the obtained Training and Validation Accuracy, it can be concluded that the model is a good fit.
- The Area Under the Receiver Operator Characteristic Curve (AUROC) shown in Fig 12 is over 0.9, which is a descent one, as more the AUROC (towards 1.0), better the performance of the model.

| Comparison Parameters | Alina Lazar [4] | Chockalingam et. al. [1] | Mohammed Topiwalla [3] | Our Model |
|---|---|---|---|---|
| Methodology | PCA and Support Vector Machine | Gradient Boosting Classifier | XGBOOST | Hyper-Parameter tuned Gradient Boosting Classifier |
| Validation Accuracy (in %) | 84.9272 | 86.29 | 87.53 | 88.16 |
| Area Under ROC Curve | 0.8911 | - | 0.9275 | 0.93 |

**Table 3. Comparison with Existing Models**

## VII. CONCLUSION

This paper proposed the application of Ensemble Learning Algorithm, Gradient Boosting Classifier with extensive Hyper-Parameter Tuning with Grid Search on Adult Census Data. Finally, the Validation Accuracy, so obtained, 88.16% which is, by the best of our knowledge, has been the highest ever numeric accuracy achieved by any Income Prediction Model so far. The future scope of this work involves achieving an over-all better set of results by using hybrid models with inclusion of Machine Learning and Deep Learning together, or by applying many other advanced preprocessing techniques without further depletion in the accuracy.